\documentclass[10pt,twocolumn,letterpaper]{article}

\usepackage[pagenumbers]{cvpr} %

\usepackage{booktabs,xcolor,colortbl,graphicx,pifont, multirow}

\definecolor{table_red}{rgb}{1, 0.7, 0.7}
\definecolor{table_orange}{rgb}{1,0.85, 0.7}
\definecolor{table_yellow}{rgb}{1,1, 0.8}

\usepackage{microtype}

\usepackage{csquotes}

\usepackage[accsupp]{axessibility}

\newcommand\blfootnote[1]{%
  \begingroup
  \renewcommand\thefootnote{}\footnote{#1}%
  \addtocounter{footnote}{-1}%
  \endgroup
}

\definecolor{cvprblue}{rgb}{0.21,0.49,0.74}

\usepackage[pagebackref,breaklinks,colorlinks,allcolors=cvprblue]{hyperref}

\def\paperID{40005} %
\def\confName{CVPR}
\def\confYear{2026}

\title{PPISP: Physically-Plausible Compensation and Control\\
of Photometric Variations in Radiance Field Reconstruction\vspace{-4mm}}

\author{
Isaac Deutsch\footnotemark[1], Nicolas Mo\"enne-Loccoz\footnotemark[1], Gavriel State, Zan Gojcic \\
\small NVIDIA \\
\small\texttt{\{ideutsch, nicolasm, gstate, zgojcic\}@nvidia.com} \\
\small\textbf{\url{https://research.nvidia.com/labs/sil/projects/ppisp/}}\\
}

\begin{document}

\twocolumn[{
\maketitle

\vspace{-11mm}
\renewcommand\twocolumn[1][]{#1}
\begin{center}
    \centering
    \includegraphics[width=1.0145\linewidth]{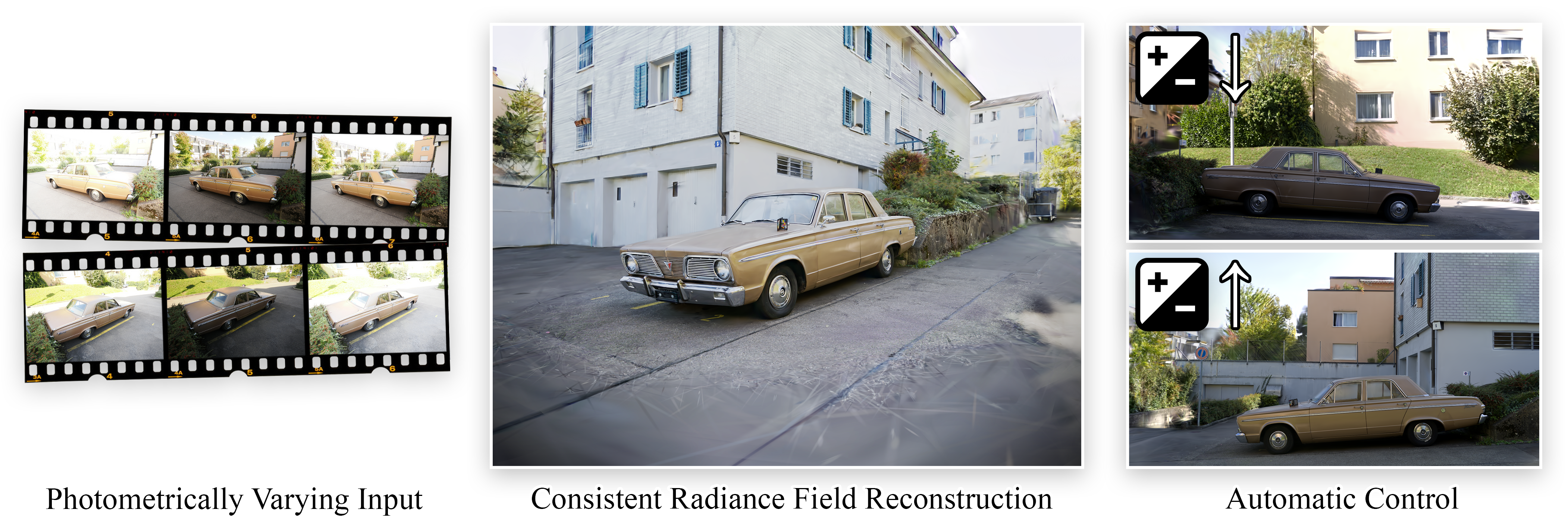}
    \captionof{figure}{We introduce a differentiable image processing pipeline applied to radiance field reconstruction. By modeling the behavior of conventional cameras, our approach disentangles image formation effects from the rest of the pipeline. Our physically-plausible model admits a controller module that predicts exposure and color changes for novel views.}
    \label{fig:splash}
\end{center}
}]

\blfootnote{\hspace{-1em} $^*$ Equal contribution.}

\vspace{-2mm}
\begin{abstract}
Multi-view 3D reconstruction methods remain highly sensitive to photometric inconsistencies arising from camera optical characteristics and variations in image signal processing (ISP). Existing mitigation strategies such as per-frame latent variables or affine color corrections lack physical grounding and generalize poorly to novel views. We propose the Physically-Plausible ISP (PPISP) correction module, which disentangles camera-intrinsic and capture-dependent effects through physically based and interpretable transformations. A dedicated PPISP controller, trained on the input views, predicts ISP parameters for novel viewpoints, analogous to auto exposure and auto white balance in real cameras. This design enables realistic and fair evaluation on novel views without access to ground-truth images. PPISP achieves state-of-the-art performance on standard benchmarks, while providing intuitive control and supporting the integration of metadata when available.
The source code is available at: \small\textbf{\url{https://github.com/nv-tlabs/ppisp}}
\end{abstract}

\vspace{-2mm}
\section{Introduction}
\label{sec:intro}

State-of-the-art multi-view 3D reconstruction methods have significantly advanced the fidelity of novel view synthesis (NVS), transforming it into a technology with real-world applications in physical AI simulation, virtual production, and content creation. Despite these advances, the quality of reconstruction and view synthesis remains highly sensitive to the quality of the input data—both to the distribution of camera poses and to multi-view appearance inconsistencies. The latter often arise from variations in camera optical characteristics and image signal processing (ISP) settings over time. These variations result in differences in color tone, intensity, and contrast that violate the photometric consistency assumptions underlying 3D reconstruction.

A common strategy to mitigate these appearance variations is to introduce additional, optimizable per-frame or per-camera parameters designed to capture photometric residuals while preserving a consistent multi-view scene representation. Recent state-of-the-art approaches include low-dimensional generative latent optimization (GLO) vectors~\cite{martinbrualla2021nerfw}, learnable affine transformations~\cite{rematas2022urf}, and bilateral grids (BilaRF)~\cite{wang2024bilateral}. However, these mitigation strategies face several trade-offs and challenges:
\begin{itemize}
    \item \textbf{Representation capacity:} higher-capacity and less-constrained modules tend to improve PSNR on the training views but risk modeling more than just photometric variations, often degrading novel view synthesis quality.
    \item \textbf{Interpretability and controllability:} the learned parameters are typically non-interpretable (\eg, in GLO or BilaRF), making it difficult to intuitively adjust properties such as brightness or white balance.
    \item \textbf{Parameters for novel views:} since the parameters are optimized independently per frame, it is unclear how to assign appropriate values when synthesizing novel views.
\end{itemize}

\noindent The latter is especially challenging due to the tendency of these modules to conflate camera sensor intrinsic properties (\eg, vignetting and camera response function) with capture-dependent settings that vary per frame or are adjusted by the ISP (\eg, exposure time and white balance). As a consequence, evaluation protocols commonly assume access to the ground-truth novel view image and estimate a corrective mapping, such as an affine transform, quadratic polynomial, or direct parameter optimization, to minimize the difference between the synthesized and the ground-truth (GT) image before computing the evaluation metrics. But such protocols are inherently flawed as they: \textbf{(i)} deviate from real-world scenarios where GT novel views are unavailable, and \textbf{(ii)} conceal differences between methods by compensating for them through the corrective mapping.

To address these challenges, we propose a Physically\mbox{-}Plausible ISP (PPISP) correction module, grounded in the physical principles of camera image formation. Specifically, we disentangle sensor-intrinsic properties and capture-dependent settings through dedicated per-sensor and per-frame modules, respectively, and constrain their effects according to the image formation process (\eg, the exposure module can only modify the overall image brightness). Our model acts as a post-processing step applied to the raw images rendered from the 3D representation, and enables direct controllability through manual change of the parameters. Moreover, we introduce a PPISP controller that predicts the parameters of the per-frame modules for novel views, analogous to the auto exposure and auto white balance mechanisms in conventional cameras.

\begin{figure*}[t]
\centering
\includegraphics[width=0.95\textwidth]{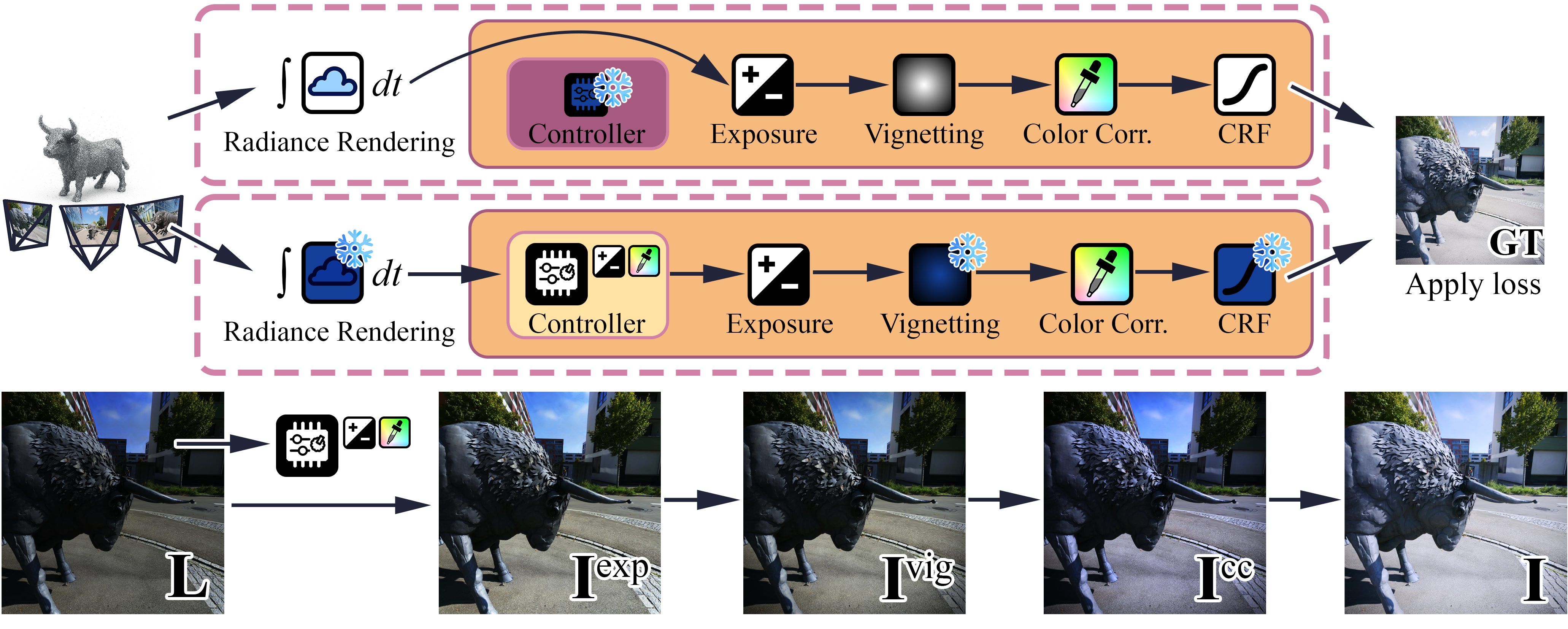}
\caption{Our proposed pipeline applies a sequence of physically-grounded modules to the input reconstructed radiance (exposure offset, chromatic vignetting, linear color correction, and non-linear camera response function). Top: all modules except the controller are jointly optimized during the first training phase. Bottom: the controller is then trained to predict per-frame exposure and color correction for novel views while other modules are frozen. The image sequence illustrates the progressive effect of each module.}
\vspace{-2mm}
\label{fig:method_overview}
\end{figure*}

\section{Related Work}
\label{sec:related_work}

Appearance inconsistencies across multi-view input images significantly degrade the quality of radiance field reconstructions and subsequent novel-view synthesis. Such variations are common in unconstrained image collections, for instance when using internet photo collections or captures under uncontrolled lighting conditions.

\vspace{-2mm}
\paragraph{Compensation during reconstruction.}
To mitigate these inconsistencies, NeRF\mbox{-}W~\cite{martinbrualla2021nerfw} and GS\mbox{-}W~\cite{zhang2024GS-W} introduce GLO that are optimized jointly with the scene representation. These per-image latent embeddings enable smooth interpolation across observed appearances, but risk entangling scene geometry with reflectance when optimized end-to-end. Block\mbox{-}NeRF~\cite{tancik2022block} extends this idea to city-scale scenes and additionally conditions on camera exposure metadata. To impose stronger constraints and better align with the image formation process, subsequent works model photometric transformations explicitly. URF~\cite{rematas2022urf} represents per-image variations using affine color transformations, while BilaRF~\cite{wang2024bilateral} extends this idea to per\mbox{-}pixel affine mappings parameterized via bilateral grids. Several works instead model specific physical components of the image formation pipeline: Xian \etal~\cite{xian2023neural} learn lens distortion and vignetting jointly with the radiance field, and HDR\mbox{-}NeRF~\cite{huang2022hdr} and HDR\mbox{-}GS~\cite{hdrgs2024} recover HDR radiance by learning a camera response function (CRF) from multi-exposure captures. Closest to our approach, ADOP's~\cite{Ruckert2022ADOP} post-processing models exposure, white balance, CRF, and vignetting effects as explicit calibration parameters. However, our formulation better disentangles exposure offset and white balance, while using a more compact CRF model. Recently, Huang \etal~\cite{ltmnerf2024} and Niemeyer \etal~\cite{nexf2025} deviate from a frame-based correction and instead learn a 3D exposure neural field, predicting the optimal exposure values for each 3D point.

\vspace{-2mm}
\paragraph{Harmonizing appearance during preprocessing.}
An alternative strategy is to decouple the compensation from reconstruction and harmonize the input images as a preprocessing step. Shin \etal~\cite{chroma2025} employ a transformer network to predict bilateral grids that harmonize each image to a chosen reference view. Alzayer \etal~\cite{alzayer2025generativemvr} instead use a diffusion model to relight images directly, but due to the lack of paired real data, they train their generative model only on synthetic data. To overcome this limitation, Trevithick \etal~\cite{trevithick2024simvs} use a generative video model to simulate capture-time inconsistencies on consistent multi-view images, creating pseudo-paired data to train a harmonization network.

\vspace{-2mm}
\paragraph{Novel view synthesis with target appearance.}
The above methods reconstruct the scene in a canonical or reference appearance, but it remains unclear how to set the parameters of their appearance modules to render an image in a desired target appearance. This target appearance could be user-defined or selected to match the appearance that a camera with auto exposure and white balance would produce. This ambiguity poses practical challenges for novel view synthesis and complicates fair evaluation under photometric variation. Prior work typically applies post\mbox{-}render normalization that assumes access to the target image during evaluation: NeRF\mbox{-}W~\cite{martinbrualla2021nerfw} fine-tunes latent embeddings on one half of each image and evaluates on the other, RawNeRF~\cite{mildenhall2022rawnerf} performs channel-wise affine alignment, Mip\mbox{-}NeRF 360~\cite{barron2022mipnerf360} uses a quadratic color basis alignment, and ADOP~\cite{Ruckert2022ADOP} re-optimizes per-frame parameters.

\noindent Such evaluation protocols, however, \textbf{(i)} mask differences between methods and \textbf{(ii)} are infeasible in real-world applications where access to the target image cannot be assumed. In line with the principle that novel views should be rendered solely from reconstructed data without access to target pixels, we introduce a PPISP controller that takes the \emph{raw} radiance image rendered from the 3D representation as input and outputs the PPISP parameters. We optimize this network on the training views and then directly apply it to the novel views during inference. Somewhat related to our PPISP controller, \cite{tomasi2021learned, onzon2021neural} train a network to predict exposure control for improved feature matching and object detection, respectively.  

\section{Preliminaries}
\label{sec:preliminaries}

\paragraph{Radiance Field Reconstruction} aims to optimize a 
parametric representation of a scene's volumetric density $\sigma \in \mathbb{R}$ and emitted radiance $\mathbf{c} \in \mathbb{R}^3$. The radiance $\mathbf{L}(\mathbf{r})$ of a camera ray $\mathbf{r}(x) = \mathbf{o} + x\,\mathbf{d}$ with origin $\mathbf{o} \in \mathbb{R}^3$ and direction $\mathbf{d} \in \mathbb{R}^3$ is rendered from this representation as

\begin{equation}
\label{eq:volume_rendering}
\mathbf{L}(\mathbf{r}) = \int_{near}^{far} T(x) \, \sigma(\mathbf{r}(x)) \, \mathbf{c}(\mathbf{r}(x)) \, dt\;,
\end{equation}
where $T(x) = \exp(-\int_{near}^{x} \sigma(\mathbf{r}(y)) \, dy)$ denotes the transmittance along the ray. The optimization is supervised using ground truth images $\mathbf{I}$ captured by one or more cameras with known intrinsics and extrinsics.
This standard formulation alone does not account for camera-specific imaging effects.

\vspace{-2mm}
\paragraph{Camera Image Formation} is the process through which the radiance $\mathbf{L}$ is converted to the final image:
\begin{equation}
\label{eq:image_formation}
\mathbf{I} = \mathcal{F}(\mathbf{L};\mathbf{\Theta})\;.
\end{equation}
Here, the function \(\mathcal{F}(\cdot)\) models the complete image acquisition process, including lens distortions (e.g., vignetting, chromatic aberrations), exposure settings (aperture, shutter time), sensor characteristics (spectral response, noise, gain), and ISP operations according to some parameters \(\mathbf{\Theta}\). While some components of this process remain constant across acquisition time, others may vary due to manual adjustments or automatic adaptation by the sensor controller.

\vspace{-2mm}
\paragraph{Notation.}
Let $\mathbf{I}\in\mathbb{R}^{H\times W\times 3}$ be an RGB image. The color at spatial location $\mathbf{u}=(i,j)$ is
$\mathbf{x}=\mathbf{I}_{i,j}\in\mathbb{R}^3$ and its $k$-th channel value is
$x=\mathbf{x}_k=\mathbf{I}_{i,j,k}\in\mathbb{R}$, $k\in\{R,G,B\}$. Operations defined on channel values or colors are understood element-wise when applied to an image.

\section{Method}
\label{sec:method}
We compensate for photometric inconsistencies across input images by jointly optimizing the scene representation together with a differentiable ISP pipeline that approximates the camera image formation function \(\mathcal{F}(\cdot)\) defined in \cref{eq:image_formation}. During optimization, this pipeline models both camera-specific and time-varying effects. During inference (\ie, when rendering novel views), the learned controller (\cref{subsec:controller}) predicts the time-varying parameters directly from the radiance $\mathbf{L}$ rendered from the scene representation.

\newcommand{\Ctrl}{\mathcal{T}}   %
\newcommand{\Vig}{\mathcal{V}}                    %
\newcommand{\Exp}{\mathcal{E}}                    %
\newcommand{\Col}{\mathcal{C}}                    %
\newcommand{\CRF}{\mathcal{G}}                    %

\noindent Our ISP pipeline consists of four sequential modules (see \cref{fig:method_overview}):    
\begin{itemize}
\item \textit{Exposure offset} accounts for aperture, shutter time and gain variations,
\item \textit{Vignetting} models optical attenuation across the sensor,
\item \textit{Color correction} models sensor spectral response and white balance adjustments,
\item \textit{Camera response function} (CRF) applies a non-linear transformation from sensor irradiance to image colors.
\end{itemize}
Following Debevec and Malik~\cite{debevec1997hdr}, the first three modules operate linearly on the scene radiance, while the CRF provides the final non-linear mapping.
\cref{fig:method_overview} puts the pipeline in the context of the radiance reconstruction and illustrates the individual parts and their effects.

\subsection{Exposure Offset}
\label{subsec:exposure}
We model exposure as a global, per\mbox{-}frame scale on the radiance using a base\mbox{-}2 exponent, mimicking photographic exposure values:
\begin{equation}
    \mathbf{I}^{\mathrm{exp}} \;=\; \Exp(\mathbf{L};\Delta t) \;=\; \mathbf{L}\,2^{\Delta t}\;,
\end{equation}
where \(\Delta t\in\mathbb{R}\) is an optimizable exposure offset. This offset represents the variation of the radiance intensity reaching the sensor and is specific to the capture. Thus, we estimate one such offset for each frame.

\subsection{Vignetting}
\label{subsec:vig}
Following Goldman~\cite{goldman2010vignette}, we model per-channel radial intensity falloff using a polynomial in the squared radius around an optimizable optical center:
\begin{equation}
\mathbf{I}^{\mathrm{vig}} \;=\; \Vig(\mathbf{I}^{\mathrm{exp}};\boldsymbol{\mu},\boldsymbol{\alpha}) \;=\; \,\mathbf{I}^{\mathrm{exp}}\cdot v(r;\boldsymbol{\alpha})\;,
\end{equation}
where \(\boldsymbol{\mu}\in\mathbb{R}^{2}\) is the optical center, \(\boldsymbol{\alpha}\in\mathbb{R}^3\) are polynomial coefficients, and \mbox{\(r=\lVert\mathbf{u}-\boldsymbol{\mu}\rVert_2\)} is the distance of the pixel location $\mathbf{u}$ to the optical center.
The attenuation factor $v(r)$ is defined as:
\begin{equation}
    v(r) \;=\; \mathrm{clip}_{(0, 1)}\!\left( 1 + \alpha_{1}\,r^2 + \alpha_{2}\,r^4 + \alpha_{3}\,r^6 \right)\;.
\end{equation}
At the start of optimization, we initialize \(\boldsymbol{\alpha}=0\) and let \(\boldsymbol{\mu}\) be the image center.

Since our vignetting model is chromatic, a falloff polynomial is defined for each color channel by distinct parameter values.

\begin{figure*}[t]
\centering
\includegraphics[width=0.95\textwidth]{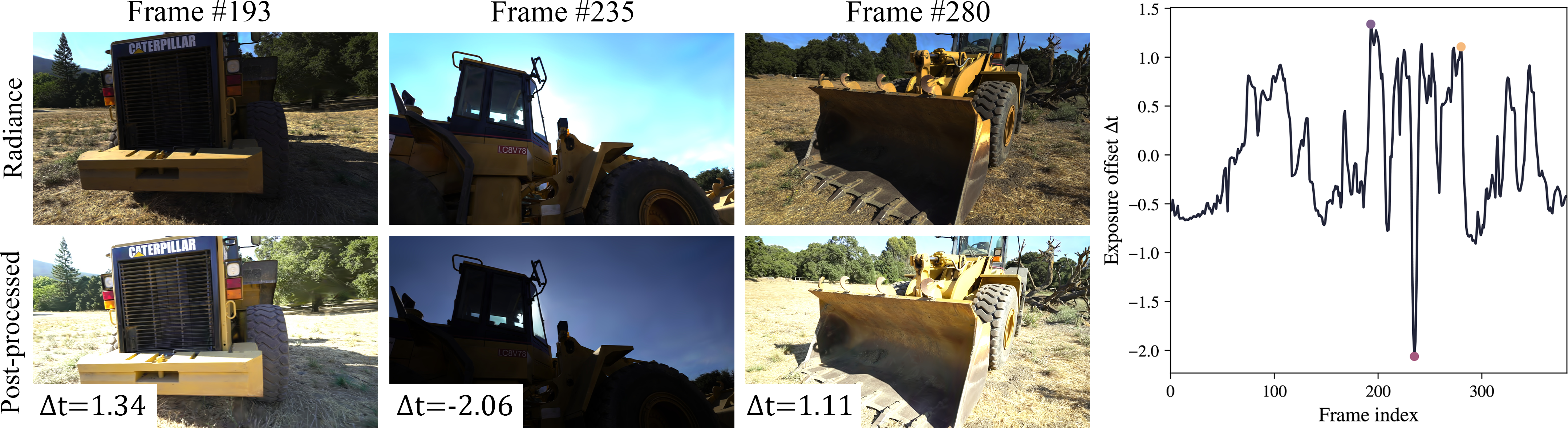}
\caption{Dynamics of the controller module. The predicted exposure offset (inset) depends on the image content of the rendered radiance. Right side: Plot of exposure offsets as predicted for each frame of the \emph{caterpillar} sequence, with the three displayed frames highlighted.}
\label{fig:controller_caterpillar}
\vspace{-2mm}
\end{figure*}

\subsection{Color Correction}
\label{subsec:cc}
To model effects such as white balance, which may vary per-frame, and gamut differences between multiple cameras, we apply color correction. To disentangle it from exposure correction, we apply a \(3\times 3\) homography \(\mathbf{H}\) on RG chromaticities and intensity --- following Finlayson \etal~\cite{finlayson2017color} --- and ensure normalization of the intensity after the transform. Inspired by DeTone \etal~\cite{detone2016deep}, we parameterize the color correction as four chromaticity offsets \(\Delta \mathbf{c}_k\), construct  \(\mathbf{H}\) from them, and apply the color correction:
\begin{equation}
    \mathbf{I}^{\mathrm{cc}} \;=\; \Col\!\big(\mathbf{I}^{\mathrm{vig}};\, \{\Delta \mathbf{c}_k\}_{k\in\{R,G,B,W\}}\big) \;=\; h(\mathbf{I}^{\mathrm{vig}};\mathbf{H})\;.
\end{equation} 
Let \(\mathbf{C}\in\mathbb{R}^{3\times 3}\) denote the RGB\(\rightarrow\)RGI conversion matrix and \(\mathbf{C}^{-1}\) its inverse. The intensity normalization can then be defined as:
\begin{equation}
    n(\mathbf{x};\mathbf{H}) \doteq \dfrac{\mathbf{x}_R+\mathbf{x}_G+\mathbf{x}_B}{\big[\mathbf{H}\cdot\mathbf{C}\,\mathbf{x}\big]_3 + \varepsilon}\;.
\end{equation}
Here, \(\varepsilon\) is a small constant for numerical stability. This normalization decouples exposure from chromatic correction. 
The color transform follows compactly as
\begin{equation}
    h(\mathbf{x};\mathbf{H}) \;\doteq\; \mathbf{C}^{-1}\!\left( n(\mathbf{x};\mathbf{H}) \cdot \big( \mathbf{H}\cdot\mathbf{C}\,\mathbf{x} \big) \right).
\end{equation}

To construct \(\mathbf{H}\), we define four 2D source–target chromaticity pairs. Specifically, we fix the source RG chromaticities $\mathbf{c}_{s,\cdot}$ to the three primaries and a neutral white:
\begin{equation}
\begin{aligned}
    \mathbf{c}_{s,R} &= (1, 0)^T\;;\quad \mathbf{c}_{s,G} = (0, 1)^T\;;\\
    \mathbf{c}_{s,B} &= (0, 0)^T\;;\quad \mathbf{c}_{s,W} = \left(\tfrac{1}{3}, \tfrac{1}{3}\right)^T\;,
\end{aligned}
\end{equation}
and define the targets $\mathbf{c}_{t,\cdot}$ as offsets from these sources $
    \mathbf{c}_{t,k} \;=\; \mathbf{c}_{s,k} + \Delta \mathbf{c}_k$ for $k \in \{R,G,B,W\}$.
By lifting the 2D chromaticities to homogeneous coordinates and stacking them as $\mathbf{S} \;\doteq\; [\, \tilde{\mathbf{c}}_{s,R}\; \tilde{\mathbf{c}}_{s,G}\; \tilde{\mathbf{c}}_{s,B}\,]$ and $\mathbf{T} \;\doteq\; [\, \tilde{\mathbf{c}}_{t,R}\; \tilde{\mathbf{c}}_{t,G}\; \tilde{\mathbf{c}}_{t,B}\,]$,
we can define
\begin{equation}
    \mathbf{M} \;\doteq\; [\tilde{\mathbf{c}}_{t,W}]_\times\,\mathbf{T}\;,
\end{equation}
where \([\cdot]_\times\) is the skew-symmetric cross-product matrix. Then, \(\mathbf{k}\in\mathbb{R}^3\) can be obtained via a cross-product of any pair of linearly independent rows \(i\) and \(j\),
\begin{equation}
    \mathbf{k} \;\propto\; \mathbf{m}_i \times \mathbf{m}_j\;.
\end{equation}
where \(\mathbf{m}_1,\mathbf{m}_2,\mathbf{m}_3\) are the rows of \(\mathbf{M}\).
Finally, we form and normalize
\begin{equation}
    \mathbf{H} \;=\; \mathbf{T}\,\mathrm{diag}(\mathbf{k})\,\mathbf{S}^{-1}\;,\qquad \mathbf{H}\;\leftarrow\; \frac{\mathbf{H}}{[\mathbf{H}]_{3,3}}\;.
\end{equation}
A precise derivation and further details are provided in the Supplementary.

\subsection{Camera Response Function}
\label{subsec:crf}
Inspired by Grossberg and Nayar~\cite{grossberg2004modeling}, we use a piecewise power curve to model non-linear chromatic transformations. The CRF operator \(\CRF\) has four learned parameters:
\begin{equation}
    \mathbf{I} \;=\; \CRF(\mathbf{I}^{\mathrm{cc}};\tau,\eta,\xi,\gamma)\;.
\end{equation}
For each channel, the basic S-shaped curve is given by:
\begin{equation}
    f_0(x;\,\tau,\eta,\xi)=
\begin{cases}
    a\left(\dfrac{x}{\xi}\right)^{\tau}, & 0\le x\le \xi\;, \\[10pt]
    1 - b\left(\dfrac{1-x}{1-\xi}\right)^{\eta}, & \xi < x\le 1\;,
\end{cases}
\label{eq:basic_f_pre_gamma}
\end{equation}
setting \(a\) and \(b\) to match the slope at the inflection point to ensure \(C^1\) continuity:
\begin{equation}
    a = \frac{\eta\,\xi}{\tau(1-\xi)+\eta\,\xi}\;, \qquad
    b = 1 - a\;.
\end{equation}

Finally, the CRF image operator \(\CRF\) is a composition of this S-curve with a gamma correction:
\begin{equation}
    \CRF(x;\,\tau,\eta,\xi,\gamma)= \big[f_0(x;\,\tau,\eta,\xi)\big]^{\gamma}\;.
\label{eq:basic_f}
\end{equation}

\begin{figure*}[t]
\centering
\includegraphics[width=\textwidth]{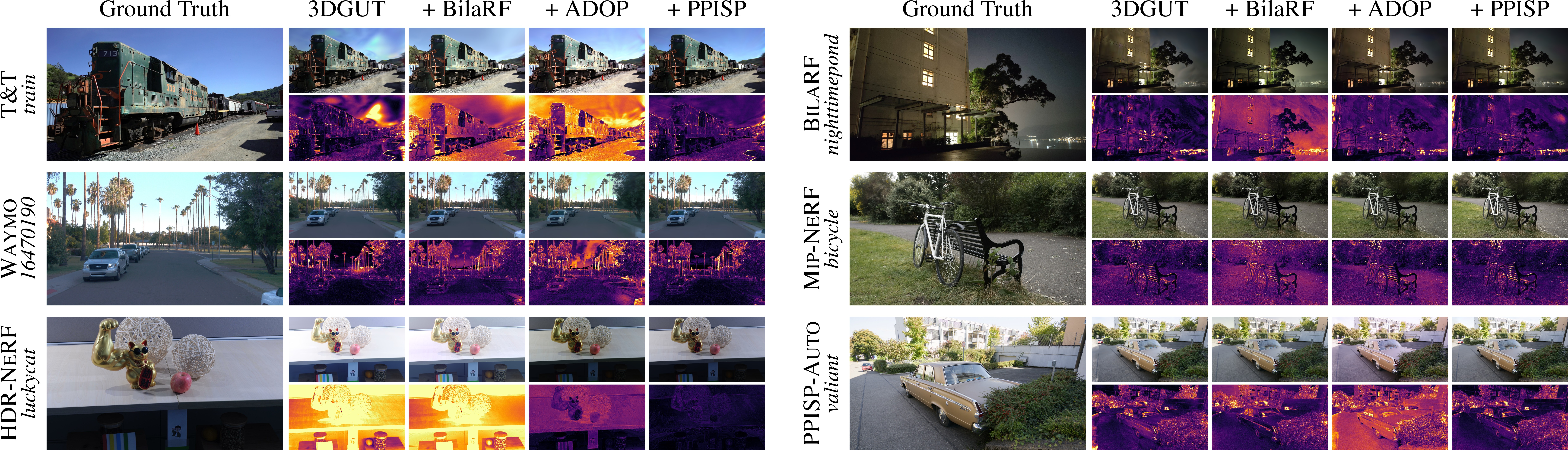}
\caption{Qualitative comparison of novel view synthesis. Row labels indicate datasets and sequences (in \textit{italics}). Column labels indicate methods. Heat maps show perceptual CIEDE2000~\cite{sharma2005ciede2000} error (colormap range: 0--20~$\Delta E_{00}$). Our method achieves more consistent photometry and better color reproduction across various datasets and sequences. HDR-NeRF: When image metadata is available, our method can incorporate it to produce a more accurate novel view.}
\vspace{-2mm}
\label{fig:qualitative_comp}
\end{figure*}

\subsection{Per-Frame ISP Parameter Controller}
\label{subsec:controller}

The exposure offsets and color correction transforms introduced above are valid only for a specific capture, \ie, a single camera pose, and therefore cannot be directly reused for novel view rendering. To address this limitation, we introduce a controller that predicts these parameters from the rendered scene radiance, analogous to how auto exposure and auto white balance work in conventional cameras:
\begin{equation}
(\Delta t,\{\Delta \mathbf{c}_k\}_{k\in\{R,G,B,W\}}) = \Ctrl(\mathbf{L})\;.
\end{equation}
Here, \(\Ctrl(\cdot)\) is the camera-specific controller parametric function, which we design as a coarse feature extractor (\(1\times1\) convolutions with pooling to a 5×5 grid), followed by a parameter regressor (an MLP with separate output heads). The detailed architecture of the controller is provided in the Supplementary.

We optimize the controller in a separate stage once the optimization of the scene representation is complete. At that stage, the underlying reconstruction and all per-camera ISP parameters are frozen, the controller-predicted parameters are applied through the ISP, and the controller itself is trained using the same photometric loss as in the initial phase. A qualitative example of the controller's effects is given in \cref{fig:controller_caterpillar}. Optional scalar controls (\eg, exposure compensation or EXIF-derived biases) can be concatenated to the regressor input.

\subsection{Regularization}
\label{subsec:regularization}
Joint optimization of the modules can introduce brightness and color ambiguities between scene radiance and the ISP parameters. To mitigate this, we apply regularization on the previously defined parameters, using the Huber loss \(\mathcal{L}_{\delta}\), where $\delta$ denotes the threshold. We use superscripts to indicate parameters belonging to specific camera sensors\(^{(s)}\) and frames\(^{(f)}\).

\vspace{-2mm}
\paragraph{Brightness.}
We penalize the mean exposure offset over frames:
\begin{equation}
    \mathcal{L}_{b} \;=\; \lambda_{b}\;
    \mathcal{L}_{\delta=0.1}\!\left(\frac{1}{F}\sum_{f=1}^{F} \Delta t^{(f)} \right)\;.
\end{equation}

\vspace{-4mm}
\paragraph{Color.}
We penalize the frame\mbox{-}mean of the target chromaticity offsets (element\mbox{-}wise in \(\mathbb{R}^2\)):
\begin{equation}
    \mathcal{L}_{c} \;=\; \lambda_{c}\sum_{k\in\{R,G,B,W\}}
    \mathcal{L}_{\delta=0.005}\!\left( \frac{1}{F}\sum_{f=1}^{F}\Delta \mathbf{c}_{k}^{(f)} \right)\;.
\end{equation}
Because chromatic corrections, as done in vignetting and CRF modules, may also introduce localized color shifts, we shrink parameter variance across channels.
Let \(\boldsymbol{\theta}_{m,k}\) be the parameters of channel \(k\) for module \(m\in\{\text{vig},\text{crf}\}\).
We penalize their across\mbox{-}channel variance, averaged over parameters:
\begin{equation}
    \mathcal{L}_{\text{var}} \;=\; \lambda_{\text{var}} \sum_{m\in\{\text{vig},\text{crf}\}} \mathrm{Var}_{k}\!\left(\boldsymbol{\theta}_{m,k}\right)\;.
\end{equation}

\vspace{-4mm}
\paragraph{Physically-plausible vignetting.}
For each polynomial, we penalize the center and softly enforce \(\alpha_{j}\le 0\):
\begin{equation}
    \mathcal{L}_{\text{vig}} \;=\;
    \lambda_{v}\!\left(\lVert \boldsymbol{\mu}_k \rVert_2^2
    \;+\; \sum_{j} \big[\alpha_{j}\big]_{+}^{2}\right)\;.
\end{equation}
Here \([x]_+=\max(x,0)\) is the elementwise rectifier.

\noindent The overall regularizer is
\begin{equation}
    \mathcal{L}_{\text{reg}}=\mathcal{L}_{b}+\mathcal{L}_{c}+\mathcal{L}_{\text{var}}+\mathcal{L}_{\text{vig}}\;.
\end{equation}

\section{Experiments}
\label{sec:experiments}
We begin by evaluating the proposed PPISP correction module and controller on standard novel-view synthesis benchmarks, assessing both reconstruction fidelity and novel-view quality (\cref{subsec:nvs_eval}). We then demonstrate how our formulation allows us to incorporate image metadata, such as relative exposure, when available (\cref{subsec:controllability}). We measure the runtime performance impact (\cref{subsec:timings}). Finally, we analyze the relationship between model capacity, overfitting behavior, and novel-view synthesis performance (\cref{subsec:overfit}). 

\vspace{-2mm}
\paragraph{Setting.} As a reconstruction-agnostic post-processing step, the PPISP module readily applies to different radiance field methods. We integrate it in 3DGUT~\cite{wu20253dgut}, GSplat~\cite{ye2025gsplat} (an accelerated implementation of 3DGS~\cite{kerbl3Dgaussians}), and Zip-NeRF~\cite{barron2023zipnerf}.

Comparison baselines are the appearance correction approaches described in GLO~\cite{martinbrualla2021nerfw}, BilaRF~\cite{wang2024bilateral}, and ADOP~\cite{Ruckert2022ADOP}. For experiments, we rely on their reference hyperparameters and reference implementations available in the respective framework. To increase the stability of ADOP's method, we increase the strength of their CRF regularization about \(100\times\) compared to the reference value. 

We jointly train the reconstruction method and the post-processing operator for 30k iterations. For the PPISP controller, we freeze both and train the controller for an additional 5k iterations. For 3DGS~\cite{kerbl3Dgaussians,ye2025gsplat} and 3DGUT~\cite{wu20253dgut}, we enable MCMC sampling~\cite{kheradmand20243d}.

\vspace{-2mm}
\paragraph{Metrics.} We evaluate the perceptual quality of the rendered views using peak signal-to-noise ratio (PSNR), structural similarity (SSIM), and learned perceptual image patch similarity (LPIPS) metrics.

As the PSNR metric is highly sensitive to global brightness shifts, and our baselines do not support appearance compensation for novel views, we additionally report the PSNR computed after affine color alignment, following \mbox{RawNeRF~\cite{mildenhall2022rawnerf}}. We denote this as \enquote{PSNR-CC}, but emphasize that such comparison masks the differences between the methods and assumes access to the GT target views, which are not available in practice.

\vspace{-2mm}
\paragraph{Datasets.}
To show the robustness and generality of our method, we conducted experiments on a variety of \mbox{publicly} available datasets: Mip-NeRF 360~\cite{barron2022mipnerf360}, Tanks and Temples~\cite{Knapitsch2017}, BilaRF~\cite{wang2024bilateral}, HDR-NeRF~\cite{huang2022hdr}, and nine static sequences of the Waymo Open Dataset~\cite{Sun_2020_CVPR}.

To further highlight the differences of the methods in challenging real-world scenarios, we captured a new \emph{PPISP dataset} consisting of four scenes. Each of them was captured with three different cameras (Apple iPhone 13 Pro, Nikon Z7, and OM System OM-1 Mark II) to ensure variations.
More details about the scenes, resolution, and training-test splits are available in the Supplementary.

\begin{table}[t]
    \centering
    \caption{\textbf{Novel view synthesis results across methods and datasets.} We compare appearance compensation methods applied on radiance field reconstruction methods. When the PPISP controller is omitted (\emph{w/o ctrl.}), novel views use zero per-frame corrections. PSNR-CC factors out global exposure and color differences.
    }
    \label{tab:results}
    \renewcommand{\arraystretch}{0.81}
    \resizebox{\columnwidth}{!}{
    \begin{tabular}{lcccc}
    \toprule
     & PSNR $\uparrow$ & PSNR-CC $\uparrow$ & SSIM $\uparrow$ & LPIPS $\downarrow$ \\
    \midrule
    \addlinespace[5pt]
    \multicolumn{5}{c}{\large\textit{\textsc{BilaRF}}} \\
    \midrule
    3DGUT~\cite{wu20253dgut}                    & 22.60 & 23.57 & 0.804 & 0.371 \\
    \quad + BilaRF~\cite{wang2024bilateral}      & 21.41 & 25.63 & 0.764 & 0.371 \\
    \quad + ADOP~\cite{Ruckert2022ADOP}          & 22.95 & 25.73 & 0.802 & 0.376 \\
    \quad + PPISP (w/o ctrl.)                    & 24.08 & 26.16 & \bf{0.820} & \bf{0.346} \\
    \quad + PPISP (w/ ctrl.)                     & \bf{24.12} & 25.92 & \bf{0.820} & 0.349 \\
    \cmidrule(lr){1-5}
    3DGS~\cite{kerbl3Dgaussians,ye2025gsplat}   & 22.55 & 24.20 & 0.799 & 0.367 \\
    \quad + BilaRF~\cite{wang2024bilateral}      & 23.06 & 25.72 & 0.766 & 0.360 \\
    \quad + GLO~\cite{martinbrualla2021nerfw}     & 21.76 & 23.28 & 0.744 & 0.398 \\
    \quad + ADOP~\cite{Ruckert2022ADOP}          & 23.42 & 26.14 & 0.804 & 0.360 \\
    \quad + PPISP (w/ ctrl.)                     & \bf{25.39} & 26.62 & \bf{0.824} & \bf{0.340} \\
    \cmidrule(lr){1-5}
    Zip-NeRF~\cite{barron2023zipnerf}           & 21.23 & 21.44 & 0.756 & 0.385 \\
    \quad + BilaRF~\cite{wang2024bilateral}      & \bf{22.94} & 24.91 & \bf{0.787} & \bf{0.336} \\
    \quad + PPISP (w/ ctrl.)                     & 22.85 & 23.76 & 0.778 & 0.368 \\
    \midrule
    \addlinespace[5pt]
    \multicolumn{5}{c}{\large\textit{\textsc{Mip-NeRF 360}}} \\
    \midrule
    3DGUT~\cite{wu20253dgut}                    & 27.74 & 27.65 & \bf{0.821} & 0.262 \\
    \quad + BilaRF~\cite{wang2024bilateral}      & 24.97 & 26.64 & 0.801 & \bf{0.260} \\
    \quad + ADOP~\cite{Ruckert2022ADOP}          & 26.42 & 27.75 & 0.815 & 0.271 \\
    \quad + PPISP (w/o ctrl.)                    & 27.55 & 28.02 & 0.819 & 0.264 \\
    \quad + PPISP (w/ ctrl.)                     & \bf{28.15} & 28.06 & \bf{0.821} & 0.264 \\
    \cmidrule(lr){1-5}
    3DGS~\cite{kerbl3Dgaussians,ye2025gsplat}   & 27.67 & 27.54 & 0.818 & 0.261 \\
    \quad + BilaRF~\cite{wang2024bilateral}      & 25.46 & 26.48 & 0.803 & 0.262 \\
    \quad + GLO~\cite{martinbrualla2021nerfw}     & 21.26 & 23.23 & 0.693 & 0.334 \\
    \quad + ADOP~\cite{Ruckert2022ADOP}          & 26.51 & 27.49 & 0.811 & 0.270 \\
    \quad + PPISP (w/ ctrl.)                     & \bf{27.97} & 27.87 & \bf{0.819} & \bf{0.260} \\
    \cmidrule(lr){1-5}
    Zip-NeRF~\cite{barron2023zipnerf}           & 27.80 & 27.64 & 0.810 & 0.257 \\
    \quad + BilaRF~\cite{wang2024bilateral}      & 26.65 & 27.05 & 0.808 & 0.255 \\
    \quad + PPISP (w/ ctrl.)                     & \bf{28.13} & 28.08 & \bf{0.815} & \bf{0.253} \\
    \midrule
    \addlinespace[5pt]
    \multicolumn{5}{c}{\large\textit{\textsc{Tanks \& Temples}}} \\
    \midrule
    3DGUT~\cite{wu20253dgut}                    & 22.86 & 23.46 & 0.790 & 0.312 \\
    \quad + BilaRF~\cite{wang2024bilateral}      & 19.78 & 23.46 & 0.770 & 0.298 \\
    \quad + ADOP~\cite{Ruckert2022ADOP}          & 20.28 & 24.20 & 0.769 & 0.323 \\
    \quad + PPISP (w/o ctrl.)                    & 21.52 & 24.87 & 0.783 & 0.296 \\
    \quad + PPISP (w/ ctrl.)                     & \bf{24.62} & 25.25 & \bf{0.809} & \bf{0.285} \\
    \cmidrule(lr){1-5}
    3DGS~\cite{kerbl3Dgaussians,ye2025gsplat}   & 23.08 & 23.73 & 0.789 & 0.303 \\
    \quad + BilaRF~\cite{wang2024bilateral}      & 20.58 & 23.34 & 0.774 & 0.296 \\
    \quad + GLO~\cite{martinbrualla2021nerfw}     & 18.74 & 20.95 & 0.707 & 0.358 \\
    \quad + ADOP~\cite{Ruckert2022ADOP}          & 20.63 & 24.28 & 0.771 & 0.314 \\
    \quad + PPISP (w/ ctrl.)                     & \bf{24.36} & 25.16 & \bf{0.807} & \bf{0.281} \\
    \cmidrule(lr){1-5}
    Zip-NeRF~\cite{barron2023zipnerf}           & 22.20 & 22.24 & 0.725 & 0.378 \\
    \quad + BilaRF~\cite{wang2024bilateral}      & 20.82 & 22.48 & 0.747 & 0.327 \\
    \quad + PPISP (w/ ctrl.)                     & \bf{23.54} & 23.85 & \bf{0.767} & \bf{0.324} \\
    \midrule
    \addlinespace[5pt]
    \multicolumn{5}{c}{\large\textit{\textsc{Waymo}}} \\
    \midrule
    3DGUT~\cite{wu20253dgut}                    & 25.56 & 25.21 & 0.785 & 0.397 \\
    \quad + BilaRF~\cite{wang2024bilateral}      & 21.83 & 23.66 & 0.768 & 0.397 \\
    \quad + ADOP~\cite{Ruckert2022ADOP}          & 24.28 & 25.18 & 0.781 & 0.405 \\
    \quad + PPISP (w/o ctrl.)                    & 25.03 & 25.46 & 0.786 & \bf{0.391} \\
    \quad + PPISP (w/ ctrl.)                     & \bf{25.69} & 25.48 & \bf{0.787} & \bf{0.391} \\
    \midrule
    \addlinespace[5pt]
    \multicolumn{5}{c}{\large\textit{\textsc{PPISP-Auto}}} \\
    \midrule
    3DGUT~\cite{wu20253dgut}                    & 22.05 & 22.20 & 0.677 & 0.453 \\
    \quad + BilaRF~\cite{wang2024bilateral}      & 20.81 & 22.30 & 0.668 & 0.440 \\
    \quad + ADOP~\cite{Ruckert2022ADOP}          & 19.94 & 22.52 & 0.670 & 0.462 \\
    \quad + PPISP (w/o ctrl.)                    & 21.07 & 23.14 & 0.677 & 0.438 \\
    \quad + PPISP (w/ ctrl.)                     & \bf{22.87} & 23.21 & \bf{0.687} & \bf{0.434} \\
    \cmidrule(lr){1-5}
    3DGS~\cite{kerbl3Dgaussians,ye2025gsplat}   & 22.29 & 22.42 & 0.679 & 0.442 \\
    \quad + BilaRF~\cite{wang2024bilateral}      & 20.92 & 22.37 & 0.668 & 0.438 \\
    \quad + GLO~\cite{martinbrualla2021nerfw}     & 19.62 & 20.53 & 0.613 & 0.488 \\
    \quad + ADOP~\cite{Ruckert2022ADOP}          & 20.22 & 22.49 & 0.670 & 0.449 \\
    \quad + PPISP (w/ ctrl.)                     & \bf{22.86} & 23.17 & \bf{0.687} & \bf{0.426} \\
    \bottomrule
    \end{tabular}
    }
\end{table}

\begin{table}[t]
\centering
\caption{Component ablation of PPISP on the Tanks and Temples dataset for novel views (NV). Each row shows performance when removing the specified component.}
\small
\begin{tabular}{l c}
\toprule
 & NV PSNR~$\uparrow$ \\
\midrule
PPISP (full)            & 24.62\\
\midrule
PPISP - no exposure & 23.33     \\
PPISP - no vignetting   & 24.08     \\
PPISP - no color correction &    24.27  \\
PPISP - no CRF  & 24.36     \\
\bottomrule
\end{tabular}
\label{tab:ablation}
\end{table}

\subsection{Novel View Synthesis Benchmark}
\label{subsec:nvs_eval}
Quantitative results on the standard benchmark scenes are presented in \cref{tab:results}, and qualitative comparisons are shown in \cref{fig:qualitative_comp}. Our method achieves the best PSNR, SSIM, and LPIPS in the large majority of settings across datasets and base methods, and on most datasets even surpasses the BilaRF baseline~\cite{wang2024bilateral} when that baseline is given privileged access to the target image, i.e., when comparing our PSNR against the baseline's PSNR-CC. These gains, established primarily on 3DGUT~\cite{wu20253dgut}, extend to the 3DGS~\cite{kerbl3Dgaussians,ye2025gsplat} and Zip-NeRF~\cite{barron2023zipnerf} integrations.

The comparison between PSNR and PSNR-CC further highlights the effectiveness of our controller in reproducing the camera’s auto-exposure and white-balance behavior. On most datasets, the controller achieves metrics close to those obtained after affine color alignment, indicating that it faithfully predicts the necessary per-frame appearance corrections. The only notable discrepancy appears on the \mbox{BilaRF} dataset, likely due to the fact that this dataset contains some manual settings overrides (indicated by the metadata), which are not captured by our controller.

Both PPISP and ADOP~\cite{Ruckert2022ADOP} employ camera-specific components (vignetting and CRF), which generalize to novel views, leading to improved metrics over BilaRF~\cite{wang2024bilateral}. Our base image formation model (\emph{w/o ctrl.}) outperforms both of these baselines thanks to better separation of concerns of the individual modules and stronger constraints (see also \cref{subsec:overfit}; a detailed comparison to ADOP is provided in the Supplementary). Our full pipeline consistently improves upon the base model by providing plausible per-frame parameter estimates via the controller.

\vspace{-2mm}
\paragraph{Ablation.} We ablate the relative contribution of each module in our pipeline through an ablation study on the \textsc{Tanks and Temples} dataset. \cref{tab:ablation} presents the novel view PSNR when individual components are removed from the full pipeline. The results demonstrate that all modules contribute to the full pipeline's performance, with exposure and vignetting corrections being most critical.

\newcommand{\cmark}{\ding{51}} %
\begin{table}[t]
\centering
\caption{Novel View PSNR across datasets with metadata. Our pipeline is able to leverage metadata (e.g. EXIF) from the sensor as a side data provided to the controller regressor.}
\label{tab:controllability}
\small
\setlength{\tabcolsep}{4pt}  %
\resizebox{\columnwidth}{!}{
\begin{tabular}{lccccc}
\toprule
& \multicolumn{1}{c}{} & \multicolumn{2}{c}{\textsc{HDR-NeRF~\cite{huang2022hdr}}} & \multicolumn{2}{c}{\textsc{PPISP}} \\
\cmidrule(lr){3-4} \cmidrule(lr){5-6}
 & {metadata} & PSNR~$\uparrow$ & PSNR-CC~$\uparrow$ & PSNR~$\uparrow$ & PSNR-CC~$\uparrow$ \\
\midrule
3DGUT~\cite{wu20253dgut}                                    & \phantom{\cmark} & 17.81 & 27.37 & 12.44 & 18.59 \\
3DGUT + BilaRF~\cite{wang2024bilateral}                     & \phantom{\cmark} & 15.40 & 26.95 & 13.39 & 20.89 \\
\cmidrule(lr){1-6}
\multirow{2}{*}{3DGUT + ADOP~\cite{Ruckert2022ADOP}}       & \phantom{\cmark} & 15.49 & 24.14 & 13.36 & 17.34 \\
                                                            & \cmark            & 31.27 & 36.10 & 20.44 & 21.60 \\
\cmidrule(lr){1-6}
\multirow{2}{*}{\textsc{3DGUT + PPISP}}                    & \phantom{\cmark} & 17.86 & 27.78 & 14.69 & 21.19 \\
                                                            & \cmark            & \textbf{34.30} & 37.10 & \textbf{21.69} & 21.94 \\
\bottomrule
\end{tabular}}
\end{table}

\subsection{Using Image Metadata}
\label{subsec:controllability}
Because our formulation closely mirrors the camera image formation process, it can naturally incorporate image metadata, such as the relative exposure of each frame, whenever available. We demonstrate this capability on the HDR\mbox{-}NeRF~\cite{huang2022hdr} and PPISP datasets, both of which use exposure bracketing (\ie, captures with positive and negative exposure compensation) and provide the corresponding metadata. We concatenate this metadata to the input of the controller MLP regressor, allowing it to map rendered radiance plus metadata to effective ISP parameters.

Since the ADOP-style post-processing also models per-frame exposure offsets explicitly, we initialize them from known exposure values as proposed in ADOP~\cite{Ruckert2022ADOP}.

Quantitative results in \cref{tab:controllability} show that supplying calibrated exposure offsets substantially improves novel-view accuracy. Moreover, providing this metadata to the controller yields further gains compared to ADOP, demonstrating our method's ability to leverage metadata for more accurate novel view appearance prediction.

\begin{table}[t]
\centering
\caption{Rendering times (ms) on NVIDIA RTX 5090 for the \mbox{MipNeRF 360~\cite{barron2022mipnerf360}} dataset.}
\small
\begin{tabular}{l c c}
\toprule
 & Time (ms)~$\downarrow$ & \% overhead~$\downarrow$\\
\midrule
3DGUT~\cite{wu20253dgut}                 & 3.24 & ---\\
\midrule
BilaRF~\cite{wang2024bilateral}                & 1.17 & 36\%\\
ADOP                  & 0.10 & 3\%\\
PPISP (w/o ctrl.)     & 0.10 & 3\%\\
PPISP (w/ ctrl.)      & 0.84 & 26\%\\
\bottomrule
\end{tabular}
\label{tab:timings}
\end{table}

\begin{table}[t]
\centering
\caption{Average PSNR on the Tanks and Temples dataset comparing training views (TV) and novel views (NV) for ISP modules with varying capacity. The limited capacity of our proposed pipeline reduces overfitting and leads to better generalization.}
\small
\begin{tabular}{l c c}
\toprule
 & TV PSNR~$\uparrow$ & NV PSNR~$\uparrow$ \\
\midrule
BilaRF + PC                              & 26.83          & 21.80     \\
PPISP + BilaRF~\cite{wang2024bilateral}  & 26.66          & 23.52     \\
BilaRF~\cite{wang2024bilateral}          & \textbf{26.87} & 19.78    \\
ADOP~\cite{Ruckert2022ADOP}              & 26.08          & 20.28    \\
PPISP                                    & 25.85          & \textbf{24.62}     \\
\bottomrule
\end{tabular}
\label{tab:overfitting}
\end{table}

\subsection{Runtime Performance}
\label{subsec:timings}
\cref{tab:timings} presents the computational performance of the post-processing methods we evaluated compared to the scene rendering. 
PPISP (\emph{w/o ctrl.}) and ADOP~\cite{Ruckert2022ADOP} have a similar and very small computational footprint ($3\%$ of the rendering). The controller is adding a substantial overhead due to the required processing of the input image, but our pipeline remains significantly faster ($26\%$ vs $36\%$) compared to BilaRF on an NVIDIA RTX 5090 GPU.

\subsection{ISP Capacity vs. Training and Novel Views}
\label{subsec:overfit}

Next, we investigate how the capacity of the correction module affects the overfitting (difference between the PSNR on training and novel views) and generalization to novel views. The bilateral grids used in BilaRF~\cite{wang2024bilateral} provide a highly expressive mechanism for modeling image operations~\cite{chen2016bilateral} extending beyond simple compensation of photometric inconsistencies. In BilaRF~\cite{wang2024bilateral}, this operation is learned independently for each frame, providing a high modeling capacity. In contrast, our PPISP module intentionally has limited capacity to prevent overfitting, but in turn cannot model complex image operations that mix spatial and intensity effects such as localized tone-mapping.

In \cref{tab:overfitting}, we therefore study hybrids of the two approaches. Adding more capacity to per-frame BilaRF~\cite{wang2024bilateral} with additional per-camera bilateral grids (+PC) does not meaningfully change PSNR on the training views as the model already has sufficient capacity. However, it does slightly improve the generalization as per-camera corrections carry over to novel viewpoints. Increasing our method’s capacity by adding per-frame bilateral grids boosts PSNR on the training views, but noticeably degrades performance on novel views due to overfitting. Overall, our formulation achieves a favorable balance between capacity and generalization to unseen views.

\section{Conclusion and Limitations}
\label{sec:conclusion}
Accurately reconstructing the radiance field of a scene requires accounting for variations in the camera imaging pipeline across the input frames. Ignoring these variations introduces strong biases, leading to spurious color shifts and geometric artifacts. In this work, we introduced a differentiable post-processing pipeline whose design permits simulating the imaging process while remaining highly constrained to prevent reconstruction bias. We further proposed a controller that improves generalization to novel views by predicting per-frame imaging parameters directly from the rendered radiance.

\vspace{-2mm}
\paragraph{Limitations.} Our method shows superior generalization to novel views (\cref{tab:results}), but it sometimes struggles to match the baselines on the training views (\cref{tab:overfitting}). This can be partially attributed to overfitting, but our formulation also ignores some important optical effects such as localized tone-mapping commonly found in modern phone cameras; lens flares, which are prominent in night scenes; and similar spatially-varying effects. While the proposed controller enables generalization to novel views, its ability to infer exposure and color-correction parameters from rendered radiance depends on the existence of meaningful correlations in the data. When such correlations are absent, for example when the physical camera controls (\eg, shutter, aperture, ISO) are manually overridden, the controller must rely on extra metadata to predict correct values.

\section{Acknowledgments}
We thank our colleagues Qi Wu, Ruilong Li, Janick Martinez Esturo, Andr\'{a}s B\'{o}dis-Szomor\'{u}, and Nick Schneider for their suggestions, feedback, and valuable discussions.

{
    \small
    \bibliographystyle{ieeenat_fullname}
    \bibliography{main}
}

\clearpage
\setcounter{page}{1}
\maketitlesupplementary
\appendix

This supplementary material provides additional experiments, method details, and implementation specifications to complement the main paper. 

\cref{sec:sup_experiments} presents extended experimental results, including a detailed comparison with ADOP's image formation model~\cite{Ruckert2022ADOP} and additional experiments on components (camera calibration and exposure identifiability). 

\cref{sec:sup_method} provides further method details, \ie, mathematical derivations of our color correction formulation and specifications of our per-frame controller architecture.

\cref{sec:sup_implementation} details optimization settings, regularization weights, learning rate schedules, and dataset specifications used throughout our experiments.

Finally, \cref{sec:sup_manual} discusses interactive manual control capabilities of our method.

\section{Additional Experiments}
\label{sec:sup_experiments}

\begin{figure*}[t]
\centering
\includegraphics[width=0.95\textwidth]{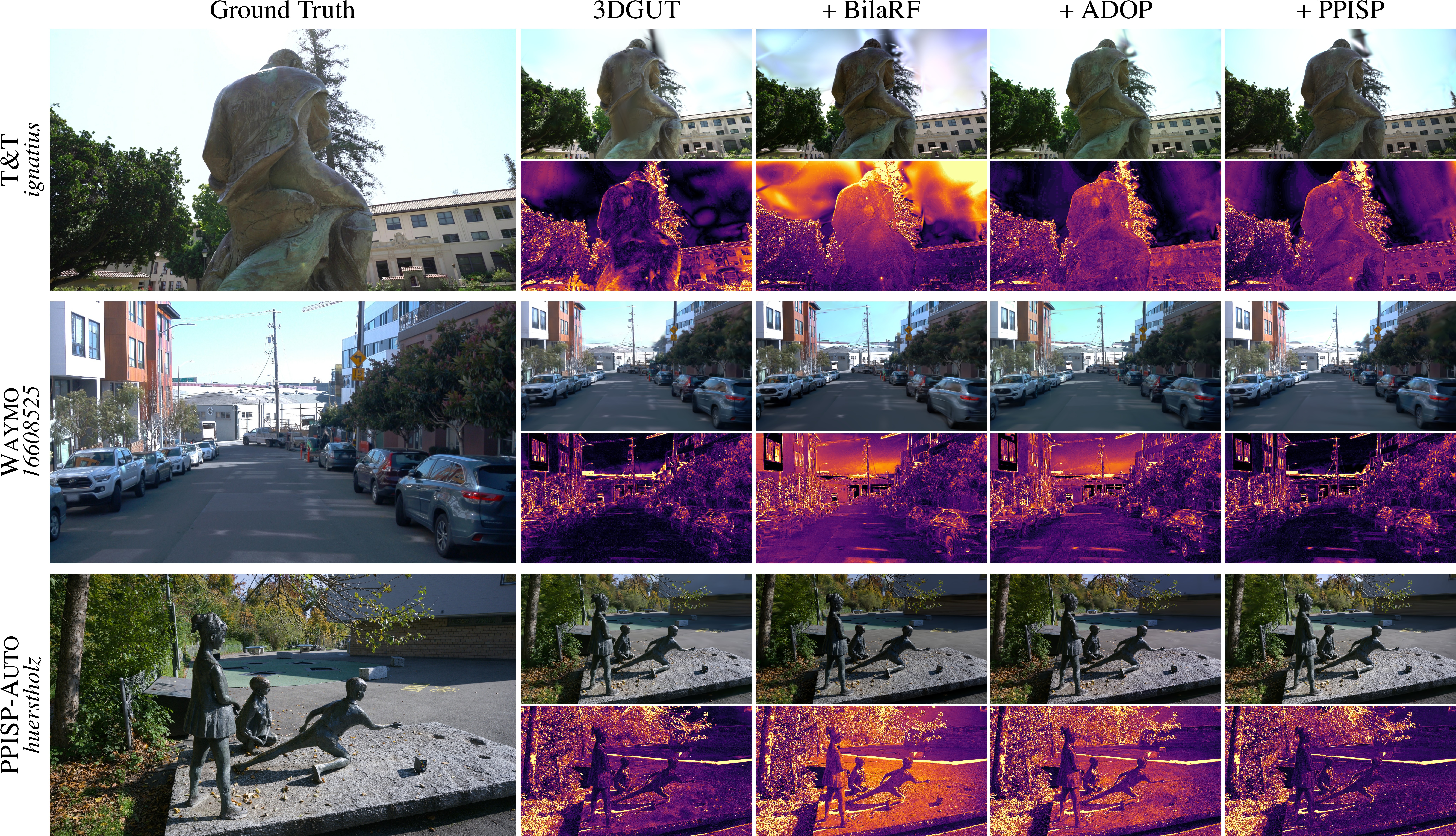}
\caption{Qualitative comparison of novel view synthesis, additional examples. Row labels indicate datasets and sequences (in \textit{italics}). Column labels indicate methods. Heat maps show perceptual CIEDE2000~\cite{sharma2005ciede2000} error (colormap range: 0--20~$\Delta E_{00}$). }
\label{fig:qualitative_comp_2}
\end{figure*}

\begin{table*}[t]
    \centering
    \caption{\textbf{Per-scene novel view PSNR comparison.} We compare post-processing methods applied on top of 3DGUT reconstruction across all sequences. Higher is better ($\uparrow$).}
    \label{tab:sup_per_scene_results}
    \renewcommand{\arraystretch}{0.9}
    \setlength{\tabcolsep}{3pt}
    \resizebox{\textwidth}{!}{
    \begin{tabular}{llcccccc}
    \toprule
    Dataset & Scene & 3DGUT~\cite{wu20253dgut} & + BilaRF~\cite{wang2024bilateral} & + ADOP~\cite{Ruckert2022ADOP} & + PPISP (w/o ctrl.) & + PPISP (w/ ctrl.) \\
    \midrule
    \multicolumn{7}{c}{\textit{\textsc{BilaRF}}} \\
    \midrule
     & building & 24.85 & 22.81 & 25.30 & 26.36 & \bf{26.46} \\
     & chinesearch & 18.34 & 20.44 & 21.27 & \bf{22.13} & 21.62 \\
     & lionpavilion & 24.16 & 24.11 & 22.89 & \bf{25.06} & 24.76 \\
     & nighttimepond & 27.11 & 21.54 & 25.07 & 27.68 & \bf{28.16} \\
     & pondbike & 25.28 & 21.17 & 24.96 & \bf{26.33} & 26.04 \\
     & statue & 22.40 & 21.01 & \bf{22.84} & \bf{22.84} & 22.26 \\
     & strat & 16.06 & 18.76 & 18.34 & 18.17 & \bf{19.55} \\
    \midrule
    \multicolumn{7}{c}{\textit{\textsc{Mip-NeRF 360}}} \\
    \midrule
     & bicycle & 25.28 & 24.26 & 24.54 & 24.95 & \bf{25.72} \\
     & bonsai & 32.52 & 28.57 & 30.33 & 32.10 & \bf{33.02} \\
     & counter & 29.36 & 26.30 & 27.58 & 28.89 & \bf{29.50} \\
     & flowers & 21.80 & 20.10 & 21.54 & 21.76 & \bf{21.95} \\
     & garden & 26.85 & 24.06 & 26.10 & 27.14 & \bf{27.31} \\
     & kitchen & 31.86 & 27.50 & 28.08 & 30.51 & \bf{32.14} \\
     & room & 32.11 & 29.53 & 30.76 & 32.95 & \bf{32.84} \\
     & stump & 26.90 & 24.90 & 26.59 & 27.03 & \bf{27.28} \\
     & treehill & 22.97 & 19.46 & 22.25 & 22.59 & \bf{23.55} \\
    \midrule
    \multicolumn{7}{c}{\textit{\textsc{Tanks and Temples}}} \\
    \midrule
     & caterpillar & 22.61 & 19.19 & 18.15 & 19.74 & \bf{25.18} \\
     & ignatius & 22.03 & 20.01 & 20.47 & 20.77 & \bf{24.04} \\
     & train & 22.06 & 19.04 & 18.95 & 20.17 & \bf{23.74} \\
     & truck & 24.72 & 20.88 & 23.56 & 25.38 & \bf{25.51} \\
    \midrule
    \multicolumn{7}{c}{\textit{\textsc{Waymo}}} \\
    \midrule
     & {\small 10275144660749673822\_5755\_561\_5775\_561} & 24.73 & 20.59 & 23.68 & 24.30 & \bf{25.17} \\
     & {\small 1265122081809781363\_2879\_530\_2899\_530} & \bf{28.39} & 24.47 & 26.30 & 27.50 & 28.31 \\
     & {\small 15959580576639476066\_5087\_580\_5107\_580} & 27.52 & 24.06 & 26.54 & 27.04 & \bf{27.77} \\
     & {\small 16470190748368943792\_4369\_490\_4389\_490} & 23.82 & 20.17 & 22.09 & 23.69 & \bf{24.21} \\
     & {\small 16608525782988721413\_100\_000\_120\_000} & \bf{23.29} & 19.86 & 22.62 & 22.91 & 23.27 \\
     & {\small 16646360389507147817\_3320\_000\_3340\_000} & \bf{26.65} & 23.71 & 24.84 & 25.86 & 26.48 \\
     & {\small 17244566492658384963\_2540\_000\_2560\_000} & 27.25 & 22.19 & 26.00 & 26.31 & \bf{27.39} \\
     & {\small 1999080374382764042\_7094\_100\_7114\_100} & 24.10 & 20.85 & 23.18 & 23.65 & \bf{24.34} \\
     & {\small 744006317457557752\_2080\_000\_2100\_000} & 24.26 & 20.53 & 23.31 & 24.05 & \bf{24.30} \\
    \midrule
    \multicolumn{7}{c}{\textit{\textsc{PPISP-auto}}} \\
    \midrule
     & huerstholz\_auto & 19.23 & 18.76 & 18.88 & 19.24 & \bf{19.81} \\
     & struktur28\_auto & 24.21 & 22.80 & 21.97 & 22.25 & \bf{25.28} \\
     & toro\_auto & 22.24 & 20.56 & 18.44 & 20.20 & \bf{23.01} \\
     & valiant\_auto & 22.51 & 21.14 & 20.47 & 22.58 & \bf{23.39} \\
    \bottomrule
    \end{tabular}
    }
\end{table*}

To complete the main paper experiments, we provide further qualitative results in \cref{fig:qualitative_comp_2} and present the detail of the novel-view PSNR for every scene in \cref{tab:sup_per_scene_results}.

\begin{figure}[t]
\centering
\includegraphics[width=\linewidth]{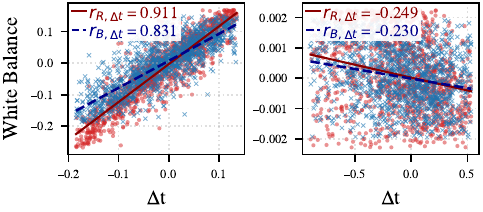}
\caption{Correlation between optimized exposure offset and white balancing variables in SMERF's~\cite{duckworth2023smerf} \emph{alameda} sequence. Left: ADOP's~\cite{Ruckert2022ADOP} red and blue channel scaling. Right: The offsets of the white point of our homography\mbox{-}based correction. The Pearson correlation coefficient for each component is inset.}
\label{fig:color_correlation}
\end{figure}

\subsection{Detailed Comparison with ADOP~\cite{Ruckert2022ADOP}}

In the related work (\cref{sec:related_work}), we mention that ADOP~\cite{Ruckert2022ADOP} implements a similar image formation model as ours. We deviate in the color correction and CRF. Here, we provide a detailed comparison, expanding on the main results in \cref{sec:experiments}.

\paragraph{White balance and exposure decoupling.}
In \cref{subsec:cc}, we claim that our color correction method, which operates on 2D chromaticities instead of 3D color and normalizes the intensity post-transformation, decouples the white balance from the exposure correction. We evaluate this by computing the Pearson correlation coefficient (PCC) between the estimated exposure offset and the white point offset, \(\Delta \mathbf{c}_W\), which controls the white balance and compare our method against ADOP's which uses per\mbox{-}channel white\mbox{-}point gains.

The PCC is defined as:
\begin{equation}
r_{X,Y} = \frac{\text{cov}(X,Y)}{\sigma_X \sigma_Y}
\end{equation}
where $r_{X,Y}$ is the Pearson correlation coefficient between variables $X$ and $Y$, $\text{cov}(X,Y)$ is the covariance between $X$ and $Y$, and $\sigma_X$ and $\sigma_Y$ are the standard deviations of $X$ and $Y$, respectively. A PCC near \(1\) indicates strong linear correlation, and a PCC near \(0\) indicates weak or no correlation.

A representative result is shown in ~\cref{fig:color_correlation}. We find that the PCC numbers for our method are substantially lower as compared to ADOP's method on all sequences, indicating an improved decoupling of white balance and exposure correction.

\begin{figure*}[t]
\centering
\includegraphics[width=0.95\textwidth]{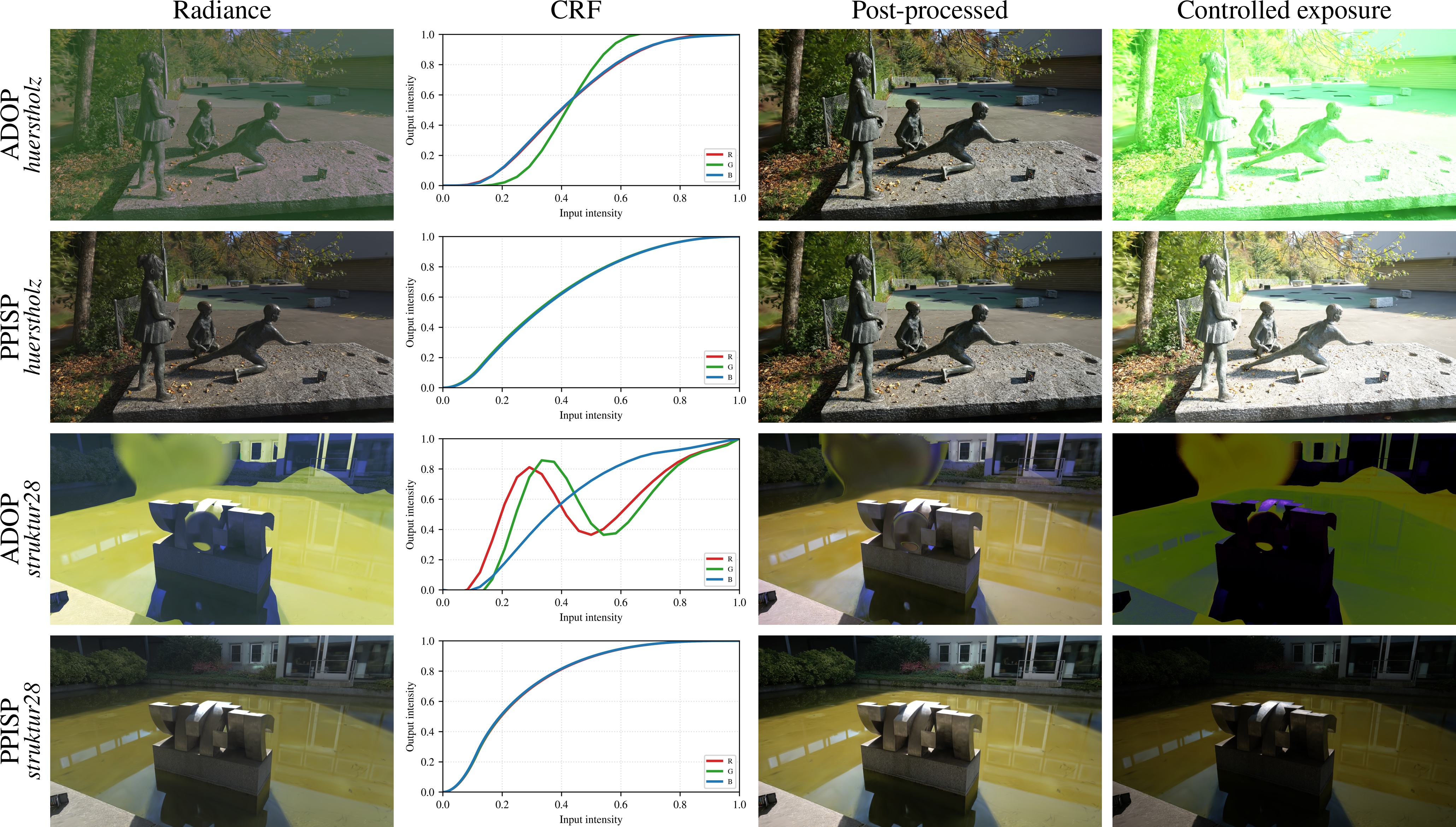}
\caption{Comparison of ADOP~\cite{Ruckert2022ADOP}-style post-processing including exposure control against our method. Row labels indicate the post-processing method and the sequence name (in \textit{italics}). The CRF for ADOP's formulation compensates for the color artifacts baked into the radiance field only at a specific exposure value. But when controlling exposure for novel views, color artifacts are exacerbated. In contrast, both our method's radiance field and output remain neutral since all corrections are decoupled.}
\label{fig:crf_comparison}
\end{figure*}

\Cref{fig:crf_comparison} further highlights the importance of decoupling color and exposure corrections: When exposure and color are coupled, the CRF will also be entangled in order to compensate for the value-dependent color shift. That, in turn, hinders the controllability of both aspects since neither can be changed without also affecting the other.

\paragraph{CRF stability in challenging sequences.}
In \cref{subsec:crf}, we provide a formulation for the camera response function that is constrained to be monotonically increasing and smooth by design. This ensures that the optimization remains stable. In some sequences, particularly when large photometric variations were present, we found that this offers an improvement over ADOP's~\cite{Ruckert2022ADOP} CRF formulation, which uses \(25\) discrete nodes which are interpolated linearly and requires a smoothness loss. A degenerate case of ADOP's CRF is illustrated in \cref{fig:crf_comparison} (third row), where the learned green and red channels of the CRF are split into lower and upper sections with a reversal. This violates the assumption that the CRF is monotonically increasing. While the post-processed image still remains close in brightness and color to the actual scene due to corrections being self-consistent, it falls apart with strong color artifacts when applying a controlled exposure offset.

\subsection{Online Camera Calibration}
\begin{figure}[t]
\centering
\includegraphics[width=0.8\linewidth]{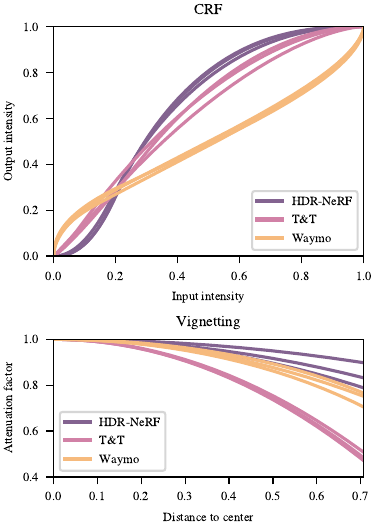}
\caption{Recovered camera-specific parameters across datasets. Top: The calibrated CRF of three sequences of each of the HDR-NeRF~\cite{huang2022hdr}, Tanks and Temples~\cite{Knapitsch2017}, and Waymo Open Drive~\cite{Sun_2020_CVPR} dataset are overlaid. Bottom: For the same sequences and datasets, the vignetting falloff curves are compared.}
\label{fig:camera_calib}
\end{figure}

Since certain parts of the PPISP pipeline, namely the vignetting (\cref{subsec:vig}) and CRF (\cref{subsec:crf}), are shared across all frames of a camera, the process of jointly optimizing them with the radiance field reconstruction can be understood as an online camera calibration. We compared the recovered per-camera parameters across multiple sequences qualitatively in \cref{fig:camera_calib}, where multiple plots are overlaid. Same color implies same dataset. The close overlap of the curves from the same datasets and the distinct shapes between datasets indicate that our method can robustly extract these calibrations. It also suggests that the camera-specific curves are disentangled from scene radiance and other corrective effects, otherwise we would expect an ambiguous mixing of them.

\subsection{Identifiability of Exposure Offsets}
In \cref{subsec:controllability}, we tested the effectiveness of using image exposure metadata to guide the image formation process. Here, we consider the inverse problem of identifying calibrated exposure offsets. In this experiment, per-frame exposure offsets are freely optimized and compared against the relative exposure metadata present in the HDR\mbox{-}NeRF~\cite{huang2022hdr} and PPISP datasets.

According to Grossberg and Nayar~\cite{grossberg2003camera}, there is an \enquote{exponential ambiguity}, which states that transforming both the inverse of the CRF and the radiance by some power produces exactly the same image intensities. Since our exposure offsets are parameterized in log-space, applying a power to the radiance corresponds to a scaling in parameter space. Thus, for this experiment, we apply an optimal affine transform on the recovered exposure offsets and compute the error on the transformed data.

\begin{figure}[t]
\centering
\includegraphics[width=0.8\linewidth]{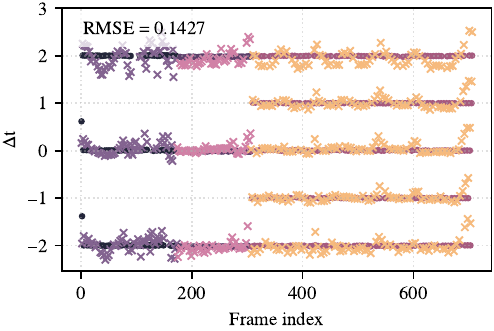}
\caption{Optimized exposure parameters per frame and given exposure metadata for the \emph{huerstholz} sequence in the PPISP dataset. Colors indicate individual cameras.}
\label{fig:identifiability}
\end{figure}

As illustrated in \cref{fig:identifiability} for a representative sequence, calibrated exposure metadata is matched closely.

\section{Additional Method Details}
\label{sec:sup_method}

\subsection{Color Correction}
In \cref{subsec:cc}, we propose a color correction method based on a \(3\times3\) homography matrix  \(\mathbf{H}\), applied on RG chromaticities and intensity, followed by an intensity normalization. For the parameterization of \(\mathbf{H}\), we show a construction from chromaticity offsets \(\Delta \mathbf{c}_k\) that control the mapping from source to target chromaticities. In this section, we provide a more detailed derivation.

Furthermore, we detail the preconditioning we apply to the chromaticity offsets \(\Delta \mathbf{c}_k\).

\paragraph{Derivation and equivalence to direct linear transformation.}
We derive the construction of \(\mathbf{H}\) in detail and show that the resulting matrix is equivalent to the standard method for constructing homography matrices from source-target pairs, the direct linear transformation (DLT).

In \cref{subsec:cc}, we define source and target chromaticity vector pairs \(\mathbf{c}_{\{s,t\},\{R,G,B,W\}}\). The homogeneous lifts of these vectors are denoted with a tilde, \(\tilde{\mathbf{c}}_{\{s,t\},\{R,G,B,W\}}\).
The \(\mathbf{S}\) and \(\mathbf{T}\) matrices are built by stacking the lifted source and target red, green, and blue chromaticity vectors, respectively. We note that \(\mathbf{S}\) is constant and has an inverse \(\mathbf{S}^{-1}\).

\paragraph{Reduction using three correspondences.}
By definition, a homography is a collinear transformation (collineation), \ie, transformed vectors are identical to the original ones up to scale: \(\mathbf{H}\,\tilde{\mathbf{c}}_{s,i}\sim \tilde{\mathbf{c}}_{t,i}\) for \(i\in\{R,G,B\}\).
Using the stacked matrices \(\mathbf{S}\) and \(\mathbf{T}\), it follows that there exist nonzero \(\mathbf{k}=(k_R,k_G,k_B)^\top\) such that
\begin{equation}
\mathbf{H}\,\mathbf{S}=\mathbf{T}\,\mathrm{diag}(\mathbf{k})
\Longrightarrow
\mathbf{H}(\mathbf{k})=\mathbf{T}\,\mathrm{diag}(\mathbf{k})\,\mathbf{S}^{-1}\;.
\end{equation}
Thus, the homography is reduced to three column scales up to a common factor.

\paragraph{Fourth correspondence via collinearity.}
To find \(\mathbf{k}\), we write the source white point as \(\tilde{\mathbf{c}}_{s,W}=\mathbf{S}\,\mathbf{b}\)
with barycentric \(\mathbf{b}=(\tfrac{1}{3},\tfrac{1}{3},\tfrac{1}{3})^\top\;\).

We require
\(\mathbf{H}\,\tilde{\mathbf{c}}_{s,W} \sim \tilde{\mathbf{c}}_{t,W}\;\). Another way to express this collinearity constraint is
\(\tilde{\mathbf{c}}_{t,W}\times \big(\mathbf{T}\,\mathrm{diag}(\mathbf{b})\,\mathbf{k}\big)=\mathbf{0}\;\).
Using the skew-symmetric matrix \([\cdot]_\times\) with \([\mathbf{x}]_\times\mathbf{y}=\mathbf{x}\times\mathbf{y}\), this yields the homogeneous linear system
\[
[\tilde{\mathbf{c}}_{t,W}]_\times\,\mathbf{T}\,\mathrm{diag}(\mathbf{b})\,\mathbf{k}=\mathbf{0}\;.
\]
For the white point, \(\mathrm{diag}(\mathbf{b})\propto \mathbf{I}\), so the constraint reduces to the \(3\times3\) system
\(\mathbf{M}\,\mathbf{k}=\mathbf{0}\;\) with \(\mathbf{M}=[\tilde{\mathbf{c}}_{t,W}]_\times\,\mathbf{T}\;\).
Generically \(\mathrm{rank}(\mathbf{M})=2\), so the right nullspace is 1D and determines \(\mathbf{k}\) up to scale. A practical closed form is to take any cross of two independent rows \(\mathbf{r}_i,\mathbf{r}_j\) of \(\mathbf{M}\), \ie:
\(\mathbf{k}\propto \mathbf{r}_i\times \mathbf{r}_j\;\).
Substituting \(\mathbf{k}\) into \(\mathbf{H}(\mathbf{k})\) and normalizing by an arbitrary scalar (\eg, set \([\mathbf{H}]_{3,3}=1\)) gives the desired homography.

\paragraph{Equivalence to the 4-point DLT.}
The classical DLT stacks the four constraints into \(\mathbf{A}\,\mathbf{h}=\mathbf{0}\) for the 9-vector \(\mathbf{h}\) of \(\mathbf{H}\) (up to scale), and solves for the 1D right-nullspace of \(\mathbf{A}\).
Our construction enforces the same constraints factorized through the invertible \(\mathbf{S}\): three correspondences reduce to the column scales \(\mathbf{k}\), and the fourth yields \(\mathbf{M}\,\mathbf{k}=\mathbf{0}\).
Under non-degenerate configurations (\ie, the columns of \(\mathbf{T}\) are not collinear and \(\mathrm{rank}(\mathbf{M})=2\)), both methods recover the same \(\mathbf{H}\) up to an overall scalar.

\paragraph{Degeneracies and identity case.}
If \(\mathrm{rank}(\mathbf{T})<2\) or \(\mathrm{rank}(\mathbf{M})<2\), \(\mathbf{k}\) is ill-defined, mirroring DLT degeneracies.
When targets equal sources, \(\mathbf{T}=\mathbf{S}\), \(\tilde{\mathbf{c}}_{t,W}=\tilde{\mathbf{c}}_{s,W}\), and \(\mathbf{k}\propto (1,1,1)\), yielding \(\mathbf{H}\) proportional to the identity after normalization.

\paragraph{Preconditioning of the chromaticity offsets.}
Our color correction method involves a conversion from RGB color to RGI (red-green chromaticity and intensity) and back, with \(I = R+G+B\) and \(B = I - R - G\) in terms of components. In our optimization setting, this correlates the gradients of the individual chromaticity offsets \(\{\Delta \mathbf{c}_i\}\) with the blue channel. In addition to that, the output image is generally more sensitive to changes in the white point than an offset in the RGB primaries.

In order to whiten the color correction and decorrelate the individual components, we apply ZCA preconditioning with proxy Jacobians following~\cite{kessy2018optimal,park2023camp}. We precondition the 8-dimensional vector of chromaticity offsets \(\{\Delta \mathbf{c}_i\}_{i\in\{R,G,B,W\}}\). We use a block decomposition into four \(2\times2\) blocks (one per control point) in place of the full \(8\times8\) transform.

\subsection{Controller Architecture}

The overall architecture of the per-frame ISP controller is given in \cref{subsec:controller}. Here, we provide the complete architectural specifications.

\vspace{-2mm}
\paragraph{Input and output.}
The controller takes as input the rendered scene radiance \(\mathbf{L} \in \mathbb{R}^{H \times W \times 3}\). Extra inputs, such as image metadata, are input at the beginning of the parameter regression stage.

The controller outputs 9 parameters: an exposure offset \(\Delta t \in \mathbb{R}\) and eight color correction offsets  \(\{\Delta \mathbf{c}_i\}_{i\in\{R,G,B,W\}}\).

\vspace{-2mm}
\paragraph{Feature extraction stage.}
The feature extractor processes the input radiance using a sequence of 1x1 convolutions and pooling operations.

\noindent
First, a 1x1 convolution maps the 3-channel input to 16 feature channels. This is followed by max pooling with a factor of 3 in each spatial dimension, reducing the resolution to \(H/3 \times W/3\). A ReLU activation is then applied. Next, a second 1x1 convolution expands the features to 32 channels, followed by ReLU. A third 1x1 convolution produces 64 feature channels, yielding a feature map \(\mathbf{F} \in \mathbb{R}^{H/3 \times W/3 \times 64}\).

Then, spatial aggregation is performed. An adaptive average pooling operation reduces the spatial dimensions to a \(5 \times 5\) grid, producing a coarse feature representation \(\mathbf{F}_{\text{pool}} \in \mathbb{R}^{5 \times 5 \times 64}\). This grid captures multi-scale spatial statistics of the scene while maintaining spatial locality, analogous to metering zones in conventional cameras.

\vspace{-2mm}
\paragraph{Parameter regression stage.}
The pooled features are flattened into a 1600-dimensional vector (\(5 \times 5 \times 64\)).
If available, image metadata may be concatenated at this stage. This is input into an MLP with three hidden layers, each containing 128 neurons with ReLU activations. The output consists of two parallel linear heads: one producing the exposure offset and the other producing the 8 color correction parameters.

\begin{figure*}[t]
\centering
\includegraphics[width=0.95\textwidth]{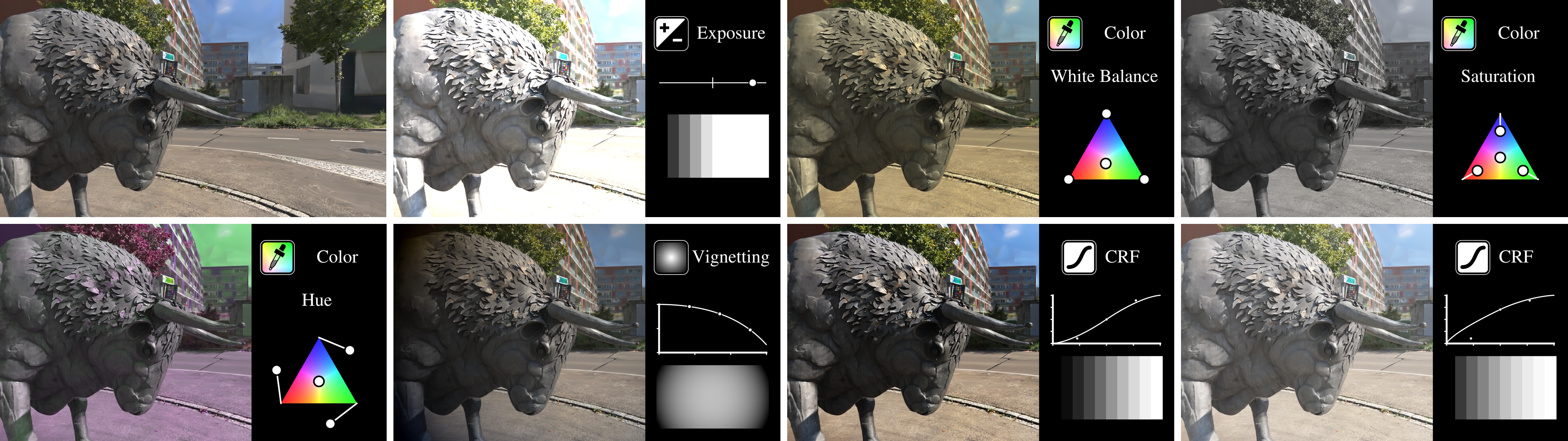}
\caption{Our low-parametric formulation of the different image processing steps enables manual editing. Top left shows the input image. Other images have details overlaid, such as the primary effect being applied and an abstract visualization. In the color correction examples, the white dots correspond to the four target chromaticities \(\mathbf{c}_{t,\{R,G,B,W\}}\), which can be intuitively manipulated.}
\label{fig:editing_gallery}
\end{figure*}

\vspace{-2mm}
\section{Additional Experiment Details}
\label{sec:sup_implementation}

We provide optimization hyperparameters, regularization weights, and dataset specifications used throughout our experiments.

\subsection{Optimization settings}

\paragraph{Regularization weights.}
In \cref{subsec:regularization}, we specify the regularizer terms that break brightness and color ambiguities and ensure physically-plausible vignetting. In \cref{sup:regularization}, we detail the numerical values used for each \(\lambda\) term.
\begin{table}[h]
\centering
\caption{Regularization coefficients.}
\label{sup:regularization}
\begin{tabular}{l c}
\toprule
\textbf{Term} & \(\boldsymbol{\lambda}\) \\
\midrule
\(\lambda_b\)        & 1.0  \\
\(\lambda_c\)        & 1.0  \\
\(\lambda_{\mathrm{var}}\) & 0.1  \\
\(\lambda_v\)        & 0.01 \\
\bottomrule
\end{tabular}
\end{table}

\vspace{-8mm}
\paragraph{Optimizer, learning rates, and schedules.}
For all post-processing modules including BilaRF~\cite{wang2024bilateral}, ADOP's formulation~\cite{Ruckert2022ADOP}, and our method, we use the Adam optimizer. We use the following learning rate scheduling with an initial delay (zero learning rate), linear warmup, and exponential decay.
\begin{equation}
lr(s) =
\begin{cases}
0, & s < s_d, \\[8pt]
lr_0 \!\left[f_s + (1 - f_s)\dfrac{s - s_d}{s_w}\right], & s_d \le s < s_d + s_w, \\[12pt]
lr_0 \,\left(f_f^{1/s_{\max}}\right)^{\,s - s_d - s_w}, & s \ge s_d + s_w .
\end{cases}
\end{equation}
Where:
\begin{itemize}
    \item $lr_0$ — base learning rate.
    \item $s$ — current training step.
    \item $s_d$ — delay steps (learning rate held at zero).
    \item $s_w$ — warmup steps (linear ramp from $f_s lr_0$ to $lr_0$).
    \item $s_{\max}$ — number of decay steps.
    \item $f_s$ — start factor for warmup (e.g., $0.01$).
    \item $f_f$ — final factor reached after decay (e.g., $0.01$).
\end{itemize}
\cref{sup:lrscheduler} details the values used during experiments.

\begin{table}[h]
\centering
\caption{Learning rate scheduler hyperparameters.}
\label{sup:lrscheduler}
\begin{tabular}{l c}
\toprule
\textbf{Term} & \textbf{Value} \\
\midrule
\(lr_0\)                & 0.002 \\
\(s_d\)                 & 0 \\
\(s_w\)                 & 500 \\
\(f_s\)                 & 0.01 \\
\(s_{\max}\)            & 30000 \\
\(f_f\)                 & 0.01 \\
\bottomrule
\end{tabular}
\end{table}

\vspace{-2mm}
\noindent
In \cref{subsec:overfit}, we experiment with combined post-processing methods. In these cases, the BilaRF module as combined with PPISP and per-camera bilateral grids uses \(s_d = 5000\) and \(s_w=1000\) with otherwise the same hyperparameters as in \cref{sup:lrscheduler}.

\subsection{Datasets}
In \cref{sec:experiments}, we outline the datasets used for experiments. In this section, we define the datasets in more detail.

\vspace{-2mm}
\paragraph{Specific choice of sequences.}
We chose the following sequences from each dataset:
\begin{itemize}
    \item Mip-NeRF 360~\cite{barron2022mipnerf360}: All nine sequences,
    \item Tanks and Temples~\cite{Knapitsch2017}: Four sequences, namely \emph{train}, \emph{truck}, \emph{caterpillar}, and \emph{ignatius},
    \item BilaRF~\cite{wang2024bilateral}: All seven sequences,
    \item HDR-NeRF~\cite{huang2022hdr}: All four real-camera sequences,
    \item Waymo Open Dataset~\cite{Sun_2020_CVPR}: Nine mostly static sequences, explicitly listed in \cref{tab:waymo_sequences}; All five cameras used.
\end{itemize}

\begin{table}[h]
\centering
\caption{Waymo Open Dataset~\cite{Sun_2020_CVPR} sequence names.}
\label{tab:waymo_sequences}
\begin{tabular}{l}
\toprule
\textbf{Sequence Name} \\
\midrule
74400631745755752\_2080\_000\_2100\_000 \\
126512208180978136\_2879\_530\_2899\_530 \\
199908037438276404\_7094\_100\_7114\_100 \\
102751446607496738\_5755\_561\_5775\_561 \\
159595805766394760\_5087\_580\_5107\_580 \\
164701907483689437\_4369\_490\_4389\_490 \\
166085257829887214\_100\_000\_120\_000 \\
166463603895071478\_3320\_000\_3340\_000 \\
172445664926583849\_2540\_000\_2560\_000 \\
\bottomrule
\end{tabular}
\end{table}

\paragraph{PPISP dataset details.}
As stated in \cref{sec:experiments}, we captured our own dataset using three cameras, including two modern mirrorless and a smartphone camera. We provide further context here.

For all cameras and scenes, we used exposure bracketing of \(\pm 2\) EV to capture HDR data. The aperture and focus were set manually and remained fixed. Image stabilization was disabled. Each scene was captured in raw format. The raw photos were developed with NX Studio and OM Workspace for the Nikon and OM System photos, and Adobe Lightroom Classic for the iPhone photos, respectively. A color calibration target placed in the scene was used to white balance.

For each scene, we additionally picked certain exposures out of the brackets and re-developed them with normalized, automatic exposure compensation and white balancing, creating a more challenging setting for the controller module. We denote this derived dataset \emph{PPISP-auto}.

\paragraph{Pre-processing.}
For all datasets including our own, where camera poses or sparse point clouds were not originally available, we processed them through COLMAP~\cite{schoenberger2016sfm} and GLOMAP~\cite{pan2024glomap} to produce the necessary inputs for the radiance field reconstruction.

We used downsampled versions of the original camera images so that the maximum effective side length of each input image did not exceed 2000 pixels. \Eg, for Mip-NeRF 360's~\cite{barron2022mipnerf360} \emph{garden} sequence, we used \(4\times\) downsampling, and for \emph{bonsai}, we used \(2\times\).

We used a seven to one split of test views to validation views for evaluation throughout.

\vspace{-2mm}
\section{Manual Control}
\label{sec:sup_manual}

Our parametric ISP formulation enables intuitive manual editing and artistic control. \cref{fig:editing_gallery} demonstrates various edits applied to a reconstructed scene, including adjustments to exposure, white balance, vignetting, and camera response. The low-dimensional and disentangled representation ensures meaningful and predictable edits, facilitating interactive workflows for applications such as artistic rendering, temporal consistency enforcement, or selective photometric matching.

\end{document}